Analyzing Examinee Comments using DistilBERT and Machine Learning to Ensure Quality

Control in Exam Content

Ye (Cheryl) Ma

## Introduction

The validity and reliability of a test are directly influenced by the quality of its items. To ensure that the items are of sufficient quality to be included in the test, multiple rounds of item review are conducted both before and after the test is administered. Typically, once the testing period has ended, psychometricians will analyze the response data using various methods to identify any items that require further review based on their statistical properties (e.g., p-value, point-biserial correlation, etc.). For example, one item with a low point-biserial correlation value can be flagged for further review due to poor discrimination.

While flagging items using their statistics can help identify potentially problematic items, it does not guarantee that the flagged items actually contain issues. Therefore, subject matter experts (SMEs) need to review the flagged items to determine whether they indeed pose any problems. Test developers can then identify the items that need to be removed from the test form/pool based on the SMEs' review results. During the review process, ways of editing items with issues can also be discussed and determined by the SMEs.

There are several reasons why items may be flagged for review. Firstly, items may be flagged based on their item statistics at both the key level and option level, such as p-value and point-biserial correlation. This can occur due to reasons such as an item being excessively easy or hard, low discrimination, or containing mis-keyed options. Additionally, items may be flagged due to issues related to model-data fit. For example, when using the Rasch model, fit is assessed using the infit and outfit statistics (Wright, Mead, & Draba, 1976). Also, various item-fit indices have been developed to assess goodness-of-fit statistics for 2PL and 3PL models, including Yen's (1981) Q1, Bock's (1972) $\chi^2$, McKinley & Mills's (1985) G2, and Orland & Thissen's (2000) likelihood-based item-fit indices. Items may also be flagged due to item parameter drift or item bias (Barba, 1977; Jones & Smith, 2006; Stahl & Muckle, 2007).

While flagging items using established approaches such as item statistics and item-fit statistics can be efficient, it lacks specific feedback regarding the issues with the flagged items. Such issues can only be identified during item review meetings. Currently, there is limited research investigating how qualitative data, such as candidates' comments, can be used to flag items. It is understandable that not all exams have the necessary features or resources for candidates to provide comments or feedback at the item level, especially for paper-based exams. However, with the increasing availability of computing power, more exams have been transferred to computer-based mode. This increases the potential for gathering item-level feedback directly from candidates during the exam, which can be a valuable source of information for flagging items.

Obtaining comments related data from candidates during the exam by allowing them to provide voluntary comments and feedback for the items can be a valuable approach. Some of these comments might provide information that can be used to identify problematic items and related issues. This method can be quicker and more direct than relying solely on statistics, which are typically obtained after the exam windows are closed. However, it is important to note that while some comments may be useful, others may not be relevant or may contain noise. A challenge arises when there are a large number of comments to review, which can be overwhelming. Reviewers must sift through the comments and decide which ones are relevant enough to be discussed during the item review meeting with SMEs. This manual review process can potentially lead to issues such as missing important negative comments or including many irrelevant comments during the item review meeting.

The practice of using customers' comments, reviews and ratings to understand customer experience and identify areas for product improvement is not new, and has been widely employed by major companies such as Amazon, Netflix, Twitter etc. This approach allows businesses to collect feedback, identify issues, and build recommendation systems. To efficiently analyze feedback from reviews in an automatic way, techniques such as sentiment analysis and topic modeling using Natural Language Processing (NLP) and machine learning (ML) have been developed and extensively researched. In recent years, NLP has gained popularity for analyzing customer and movie reviews. By using algorithms to analyze and understand human language, businesses can extract insights from large volumes of customer feedback, identifying common themes and sentiment. For example, in the movie industry, NLP techniques are used to analyze reviews and provide insights into audience reactions. By conducting sentiment analysis, companies can determine how viewers feel about a movie, which can influence future marketing and promotional efforts, ultimately improving their products and services.

**Sentiment analysis and NLP**

Sentiment analysis, also known as opinion mining, is a popular natural language processing technique that involves the identification and extraction of emotions, attitudes, and opinions expressed in text data. With the increasing availability of large amounts of data on different platforms, sentiment analysis has become a crucial tool for understanding customers' opinion and sentiment towards different products, services, and events. Compared to traditional classification machine learning models, large language models (LLM) and transformer-based models have been widely used in sentiment analysis due to their ability to generate high-quality texts and capture complex linguistic structures. For instance, in a recent study, González-Carvajal, S., & Garrido-Merchán, E. (2020) compared BERT with tradition machine learning models including voting classifier, logistic regression, linear SVC, multinomial NB, Ridge classifier and passive aggressive classifier on text classification. The study showed that BERT outperformed the traditional ML models. Similarly, Taneja & Vashishtha (2022) compared DistilBERT, BERT and traditional machine learning approach for text classification and demonstrated that DistilBERT outperforms both BERT and traditional machine learning models for various text classification application.

One of the most popular transformer-based models is the BERT model (Bidirectional Encoder Representations from Transformers). DistilBERT is a smaller and faster version of it that has been widely used in natural language processing tasks, including sentiment analysis or text classification. According to a study by Sanh et al. (2019), DistilBERT achieves a performance similar to BERT while being 60% faster and requiring only half the memory.

In summary, similar to other fields, the testing industry can leverage NLP to review and understand candidates' comments, leading to a more efficient item review and maintenance process. The purpose of this study is to explore methods for automatically identifying the most relevant candidates' comments and related items for item review using NLP and ML. Additionally, this study will investigate the efficacy of using both candidates' comments and item statistics in machine learning models. Item flagging results will be utilized as a criterion to evaluate the performance of the ML models.

This study has several research objectives, including (1) developing a model that can automatically identify the most relevant negative comments for item review, (2) exploring whether the inclusion of psychometric features in the model can enhance its performance, and (3) comparing the items flagged based on the identified comments with the items flagged in real data.

## Method

### Data

The data used in this study is from a high-demand IT certification exam administered from October 2021 to January 2022. The exam administered three non-overlapping scored forms, with a total of 150 items. Each candidate received 50 operational items and 15 pretest items. Pretest items are randomly selected from a pretest pool. The exam duration was 130 minutes, and only candidates taking English forms were included in the data analyzed for this study. Overall, 30,124 candidates completed the exam, but after data cleaning and forensics checks, the responses of 28,081 candidates were used for item parameter calibration and item statistics calculation.

The item statistics analyzed in this study for each item include Rasch-based item difficulty (b), p-value (p), point biserial correlation (r), average item response time (time), and the number of candidates to whom the items were administered after the data cleaning step or item exposure counts (n). For operational items, the magnitude of item parameter drift was also calculated. Additionally, p and r were calculated at the option level.

During the exam, candidates had the option to voluntarily provide comments for each item. The testing system allowed candidates to review and revise both their answers and the comments they provided. Therefore, the comments-related data is at the person-by-item level. Since providing comments was not mandatory, not every candidate provided comments for every item. In this dataset, there are 3,941 comments in English.

**Data labeling**

Data labeling is a crucial step for running machine learning (ML) models or fine-tuning existing language models. In this study, item analysis and item review were conducted immediately after the exam window closed. As a result, there were manual review notes from the human reviewers in the comments section, which included feedback from candidates as well as results from the item review meeting. Based on the human reviewers' notes, the author was able to identify which comments were sent to the item review meeting for further discussion. These comments were labeled as 1, while the remaining comments were labeled as 0. Also, it should be noted that the human reviewers reviewed both the comments and other psychometric information when selecting comments for review. Some comments may show negative sentiment but may not be relevant or useful enough for SMEs' review. These comments were also labeled as 0 in the dataset.

**Research design and study factors**

This study considered three main factors, including (1) whether a ML-based classification model is necessary after fine-tuning the DistilBERT model, (2) whether the inclusion of features related to item statistics and exam score can enhance model performance, and (3) if a ML-based classification model is built, which model (XGBoost or Random Forest) performs better in terms of classification accuracy and identifying a reasonable number of comments for review. Table 1 provides a summary of these study factors, as well as the research design and model conditions.

Table 1. Research design and model conditions

| Base model (BERT) | XGBoost/RandomForest (ML) |
|---|---|
| comments (text classification) | NA |
| | comment.n |
| | comment.n + item statistics + exam score |

Random Forest (James, et al, 2013) and XGBoost (Chen & Guestrin, 2016) are both widely used ensemble learning algorithms that aim to improve predictive performance by combining multiple base models. Although they share some similarities, they have distinct approaches and characteristics that can affect their performance in different contexts. Random Forest is an ensemble algorithm that builds multiple decision trees using a random subset of features and samples. Each tree is trained independently, and the final output is the majority vote or average prediction of all trees. On the other hand, XGBoost is a gradient boosting algorithm that combines weak learners (decision trees) sequentially and optimizes a differentiable loss function. It uses a regularized objective function to control model complexity and prevent overfitting. In general, both Random Forest and XGBoost have different strengths and weaknesses, and the choice between them depends on the specific characteristics of the dataset and the nature of the problems.

**Analysis procedure**

The analysis in this study consisted of three main steps (refer to Figure 1). First, the author fine-tuned the DistilBERT-base-uncased model from Hugging Face (Sanh et al., 2019) using the candidates' comments data. This provided BERT-based classification results based solely on candidates' comments, and this model is referred to as "Model 1". Additionally, the fine-tuned model provided the probability of a comment being a relevant negative comment, which was used as one of the comment-related features in the ML models.

Secondly, the author built two ML models using only comment-related features, including the probability of a comment being a relevant negative comment from Step 1 and "comment.n", which represents the number of comments an item received. The value of "comment.n" remained the same for different comments under the same item. Two popular ML models (shown in blue in Figure 1) were conducted, including XGBoost (Model 2) and Random Forest (Model 3).

Finally, to investigate whether adding features related to item statistics and exam scores could enhance model performance, the author built another two ML models using both comment-related features from Step 2 and features related to item statistics and candidates' exam scores. These included b, p, r, time, n, drift.flag, item.type, and candidates' exam score. XGBoost was used for Model 4, and Random Forest was used for Model 5 (shown in red in Figure 1).

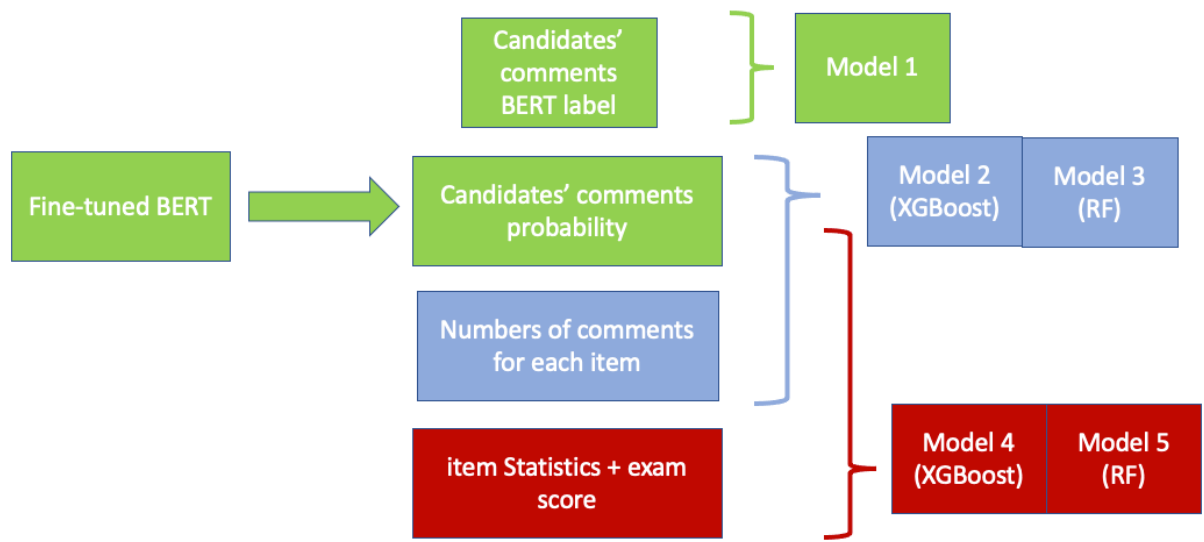

Figure 1. Research design and analysis flowchart

**Model Tuning, Training and Testing**

Two types of models are used in this study: the DistilBERT model from the transformer library and regular machine learning models. To fine-tune the DistilBERT model, the data was split into training and test sets using the typical 80 (training) -10 (5-fold cross validation) -10 (testing) split. Fine-tuning was conducted using grid searching of the number of epochs. The final model with the highest F1-score was saved to obtain BERT-based classification results and the probability of comments for the entire dataset.

For the regular ML models, the data was also split into training and test sets using the typical 80 (80% training + 20% 5-fold cross valuation) -20 (testing) split. Hyperparameters were tuned using grid search. The models were applied to both the test data and the full dataset. The parameter tuning, model training, and testing were conducted using the scikit-learn 1.1.1 Python library (Pedregosa et al., 2011).

**Evaluation Criteria**

Six evaluation metrics were computed for each model, including accuracy, false positive rate (FPR), false negative rate (FNR), precision, F1-score, and actual predictive rate. Accuracy represents the proportion of correctly classified results by the ML model. However, in cases of imbalanced data, accuracy can be misleading. Therefore, other evaluation metrics were used in this study. The false negative rate is related to recall, which is equal to 1 - false negative rate. The F1-score is the harmonic average of precision and recall, with higher values indicating better precision and recall. The actual predicted rate was also included because, in an operational setting, an ideal model should not only include all the "true positive results" but also flag a reasonable number of comments for review. For example, if a model can accurately identify all relevant negative comments by flagging all comments, the model is useless in a real setting. Therefore, generally, the higher the F1-score and the lower the actual predictive rate, the better the model.

Additionally, the number of flagged items based on the model prediction results was reported. These results were compared to the actual flagged number of items to determine whether the model could accurately flag both relevant comments and items with issues.

# Results

## Overall performance

Table 2 summarizes the components of all five models, while Tables 3 and 4 provide related false positive and false negative rates and model evaluation metrics. Among all five models, Model 4 performed the best with the lowest false negative rate and the lowest number of actual predictive comments (FP+TP). Model 5 ranked as the second-top model with respect to these metrics. Model 1 ranked the last among all models. However, when evaluating the overlap between flagged items based on the ML models and the actual flagged items, none of the models produced satisfactory results. One potential explanation could be that some problematics items might be missed when human reviewers selected the comments manually. Therefore, an iterative process and collaboration with the human reviewers are needed as the next step.

Table 2. Model Summary

| Model | Components |
|-------|-----------|
| M1 | DistilBERT |
| M2 | DistilBERT + comment.n (XGBoost) |
| M3 | DistilBERT + comments.n (Random Forest) |
| M4 | DistilBERT + comment.n + item statistics + exam score (XGBoost) |
| M5 | DistilBERT + comment.n + item statistics + exam score (Random Forest) |

## False Negative and False Positive Results

In terms of misclassified comments, all models had a small portion of comments missed from human reviewer-labeled results (false negatives) except Model 4. This indicates that the first four models missed only a small number of comments that were identified by human reviewers. Model 4 caught all the flagged comments from human reviewers. In addition, there were a fair number of comments identified by these models but not human reviewers (false positives). As mentioned in the methods section, an ideal model should not only accurately identify all the flagged comments by human reviewers but also have a small number of flagged comments from the model results (FP+TP). Therefore, among all five models, Model 4 flagged the least number of comments, whereas Model 1 flagged the highest number of comments.

Table 3. False Positive and false negative results

| Metrics | M1 | M2 | M3 | M4 | M5 |
|---------|------|-----|-----|-----|-----|
| FP | 1106 | 465 | 661 | 242 | 635 |
| FN | 7 | 2 | 18 | 0 | 15 |
| FP+TP | 1235 | 600 | 779 | 378 | 756 |

Regarding the model evaluation metrics (Table 4), Model 4 produced the highest precision, while Model 1 had the lowest precision. In terms of recall, Model 4 had the highest value, while Model 3 had the lowest. In terms of F1-score, Model 4 had the highest value, Model 5 ranked second, and Model 1 ranked last.

Table 4. Model evaluation metrics

| Metrics | M1 | M2 | M3 | M4 | M5 |
|-----------|------|------|------|------|------|
| Precision | 0.1 | 0.22 | 0.15 | 0.36 | 0.34 |
| recall | 0.95 | 0.99 | 0.87 | 1.00 | 0.90 |
| F1-score | 0.19 | 0.36 | 0.26 | 0.53 | 0.50 |

**An in-depth look of False Positive results**

Although false positive results contain the model-identified comments that were not identified by the human reviewers, it is possible that the human reviewers might have missed some important comments during the manual review. Therefore, Figure 4 was created to visualize the BERT-based probability of negative comments across all five models. Model 1 has the widest coverage, ranging from 0.5 to almost 1. Other models, especially Model 4 and Model 5, tend to have higher probabilities. Figure 5 takes a closer look at Model 4 and Model 5's results. Model 4, as the best performer among all five models, shows that most comments in the FP group have higher probabilities towards the higher end. This could be an indicator that the human reviewers might have missed some relevant comments during the manual review process.

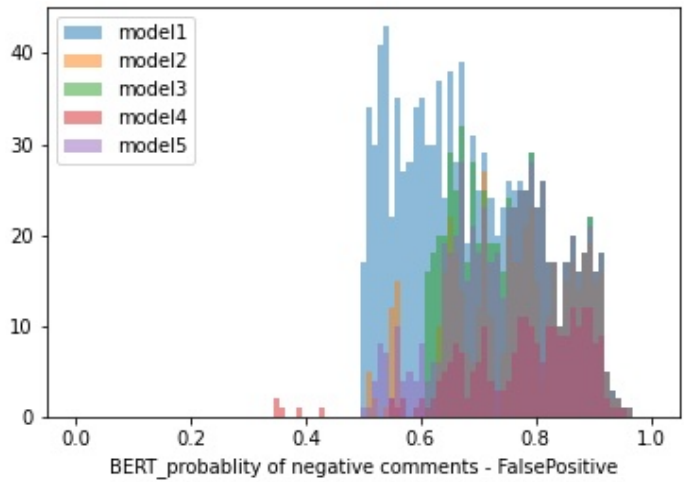

Figure 2. Distribution of BERT-based probability of negative comments

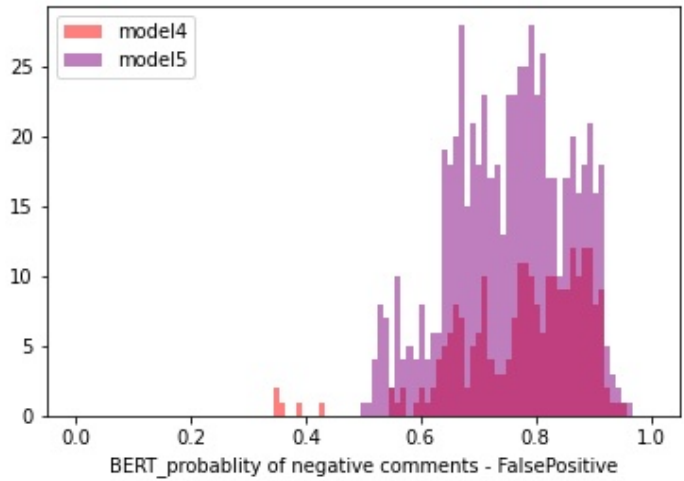

Figure 3. Model 4 and Model 5's distribution of BERT-based probability of negative comments

**Item level evaluation**

The analysis included 257 items, of which 23 were flagged for review based on what happened in the operation. The final evaluation was to show how many of the ML-flagged items overlapped with the "true" flagged items. Table 5 presents the overlapping percentage and count of items. The total flagged items column indicates the number of items flagged based on ML-flagged comments.

Model 1 flagged all of the items that were sent for review, but it also flagged 96.1% of all items, which is not practical for SMEs to review. Model 4, which performed the best according to other criteria, did not produce satisfactory results in terms of overlapping flagged items. It flagged only 36% of the items and identified only 12 out of 23 flagged items. However, for the items that were captured by the comments but not on the actual flagged items list, it may be worth reviewing these items, since those items were not reviewed by SMEs because they were not flagged by item statistical and human reviewers' comments review. It is possible that these items were missed during the initial review process.

Table 5. Flagged items' overlap between ML models vs the actual results

| Model | Flagged_items overlap N(%) | Flagged_items total N(%) |
|-------|---------------------------|--------------------------|
| M1 | 23 (100.0) | 247 (96.1) |
| M2 | 19 (82.6) | 162 (63.0) |
| M3 | 21 (91.3) | 188 (73.2) |
| M4 | 12 (52.2) | 93 (36.2) |
| M5 | 19 (82.6) | 197 (76.7) |

Notes: True results are flagged = 23, keep = 234, total = 257

## Discussion

### Overview

In summary, this study proposed an approach that automatically flags the most relevant candidates' comments for review. Results indicate the incorporating both comments' features such as sentiment scores and psychometric feature such as b, p, r can improve the model performance in identifying the most relevant comments. The findings suggest that this approach has practical implications for efficiently identifying problematic items for review. Furthermore, a phase 2 study is proposed to explore the item flagging results based on the identified comments.

**Research objective 1**

The main objective of the study was to develop a model that could automatically flag the most relevant comments for review, thus reducing the manual effort required by human reviewers to sift through all the comments to make classification decisions. Model 4 was proved to be the best performer, catching all the comments that were sent for review and flagging a reasonable number of comments (378). Therefore, by using the results of model 4, human reviewers only need to review 378 comments instead of 3875 comments, and all the "true" flagged comments are included in those 378.

**Research objective 2**

Another objective of this study was to determine whether adding psychometric-related features, including item statistics and candidates' exam scores, would improve model performance. Comparison between model 4/5 and model 2/3 indicates that adding psychometric-related features is necessary. This can be validated by human reviewers' decision-making procedure. Specifically, human reviewers take into account not only the sentiment of comments, but also item difficulty and candidates' ability when deciding whether a comment is relevant enough to send for review. For example, if a candidate has a very low exam score and provides a negative comment on a very easy item, this comment might not be worth sending for review given limited resources such as SMEs' time and workshop budgets.

**Research objective 3 and Phase 2 analysis**

The final objective of the study was to compare the items flagged based on the identified comments with the actual flagged items in the real operation setting. However, the results were not satisfied but it was not a surprise since it was anticipated due to the analysis procedure. The data labeling used in both fine-tuning the DistilBERT model and training the ML models was based on human reviewers' decisions at the comment level. Therefore, the model was trained to understand and classify comments rather than items, leading to the inability of the model to accurately identify all flagged items.

This issue highlights the need for a phase 2 study, which will focus on identifying problematic items using both comments-related and psychometric-related features at the item level. The data labeling will be based on whether an item is flagged for review in the real setting. Therefore, the phase 2 model's results will flag items instead of comments, which should address the issue of the models trained on identifying comments instead of items.

Phase 2 analysis will involve a fine-tuning process to obtain the classification results and probability of negative sentiment of comments at the item level. Instead of relying on human reviewers' notes as in the current study, obtaining the pure sentiment score can eliminate noise in the data resulting from differences among human reviewers. Additionally, other powerful LLM models will be explored and compared to obtain the optimal one that produces more accurate sentiment labels and scores for all comments.

The current study is an initial exploration, and only one exam from one testing window was used. For future operational use of Model 4, additional exams from various testing windows will be utilized to ensure that the final model is generalized enough to be applied to all exams ranging from various subjects as well as proficiency levels. Furthermore, the model will be evaluated to determine whether it can identify item issues during testing windows, particularly with pretest items, so that they can be masked if necessary.

In conclusion, this study conducted an initial exploration of using NLP and ML to automatically flag comments and items for item review. The results showed that combining qualitative features (comments) and psychometric features improved the performance of flagging relevant comments. Moreover, the study demonstrated an efficient use of language models and machine learning techniques. The next step, phase 2 analysis, will extend the analysis to automatically flag both comments and items together. This study is a promising step towards reducing the workload of product managers and improving the efficiency of the item review process.